\documentclass{article}

\usepackage{arxiv}

\usepackage[utf8]{inputenc} 
\usepackage[T1]{fontenc}    
\usepackage{hyperref}       
\usepackage{url}            
\usepackage{booktabs}       
\usepackage{amsfonts}       
\usepackage{nicefrac}       
\usepackage{microtype}      
\usepackage{lipsum}
\usepackage{graphicx}
\graphicspath{{./images/}}

\usepackage{amsmath}

\title{Financial series prediction using Attention LSTM}

\author{
  Sangyeon Kim \\
  Department of Computational Science\\
  Seoul National University\\
  \texttt{kimsasng@snu.ac.kr} \\
   \And
  Myungjoo Kang \\
  Department of Mathematics\\
  Seoul National University\\
  \texttt{mkang@snu.ac.kr} \\
}

\begin{document}
\maketitle

\begin{abstract}
Financial time series prediction, especially with machine learning techniques, is  an extensive field of study. In recent times, deep learning methods (especially time series analysis)have performed outstandingly for various industrial problems, with better prediction than machine learning methods. Moreover, many researchers have used deep learning methods to predict financial time series with various models in recent years. In this paper, we will compare various deep learning models, such as multilayer perceptron (MLP), one-dimensional convolutional neural networks (1D CNN), stacked long short-term memory (stacked LSTM), attention networks, and weighted attention networks for financial time series prediction. In particular, attention LSTM is not only used for prediction, but also for visualizing intermediate outputs to analyze the reason of prediction; therefore, we will show an example for understanding the model prediction intuitively with attention vectors. In addition, we focus on time and factors, which lead to an easy understanding of why certain trends are predicted when accessing a given time series table. We also modify the loss functions of the attention models with weighted categorical cross entropy; our proposed model produces a 0.76 hit ratio, which is superior to those of other methods for predicting the trends of the KOSPI 200.
\end{abstract}


\section{Introduction}
 Predicting the trends of financial markets is one of the most important tasks for investors. Moreover, these predictions are useful for real investing and for analyzing the general direction of other financial markets indexes.
Many have tried to predict stock market trends using methods such as technical and fundamental analysis. Technical analysis is a traditional method that uses historical stock prices and trading volumes to determine the trends of future stock movements. This method is based on supply and demand in financial markets. It facilitates the easy construction of models as it only considers numerical variables. Fundamental analysis predicts stock prices by using intrinsic values. When using this method, stock values are determined by financial news, market sentiments and economic factors; investors estimate the profits of firms and evaluate whether they are suitable for investment. 
Methodologies for forecasting stock prices have researched for a many years and several techniques have been proposed in various academic fields and applied in real markets. Quantitative methods in finance frequently use machine learning techniques.\\
These days, deep learning has been widely used in classification problems when taking advantage of nonlinearity, and has performed well compared with other classification methods. Furthermore, some research has compared deep learning with time series models for predicting time series data. For example, Kohzadi \cite{kohzadi1996comparison} compared the performance of artificial neural networks (ANN) with that of autoregressive integrated moving average (ARIMA) models for forecasting commodity prices. Kara \cite{kara2011predicting} have applied ANNs and support vector machines(SVMs) to predict prices on the Istanbul Stock Exchange(ISE) National 100 Index; the ANNs showed better accuracy than the SVMs with polynomial kernels.
 Deep learning techniques for time series data, especially those using long short-term memory (LSTM) models, have shown s results than previous machine learning techniques in speech recognition \cite{graves2013speech}, sentimental analysis \cite{palangi2016deep}, and time series prediction. Moreover, attention mechanisms are widely used for analyzing both image and time series data \cite{xu2015show}, and lead to better results when combined attention with LSTM than other plain deep learning models. Attention networks help facilitate visualization by plotting attention vector weights for understanding models.\\
 In this paper, we predict the trends of the KOSPI 200 with various deep learning models and compare their accuracy. The data set used consists of various index parameters, such as currency, global index, and commodities. We acquired training data from the beginning of 2007 to the end of 2016, and test data from the first trading day of 2017 to the end of July 2018. We achieved optimal trend prediction accuracy with weighted attention networks and illustrate the reasons for this accuracy through visualizing the attention vectors used in our work. The remainder of this paper is organized as follows:\\ In Section 2, we describe related work. Section 3 details the dataset treated in this study. Section 4 describes various methodologies used for stock prediction. Section 5 explains experimental results and their analysis. Finally, Section 6 presents our conclusions.

\section{Related works}

Kohzadi \cite{kohzadi1996comparison} tested ANN and ARIMA models for forecasting commodity prices and compared the results of each. They found that the ANNs returned 27$\%$ and 56$\%$ lower mean squared errors than the ARIMA models. Kara \cite{kara2011predicting} applied ANNs and SVMs to predict the Istanbul Stock Exchange (ISE) National 100 Index prices. Ten technical indicators were used as inputs and produced maximum values of 75.74$\%$ and 71.52$\%$ for ANNs and SVMs with polynomial kernels, respectively. The inputs were based only on the technical factors, which use historical index prices and volume data. However, the experimental procedures in  \cite{kohzadi1996comparison} and \cite{kara2011predicting} are not practical for investors because the training and test data were used without considering any time series data. Training sets should not be newer than test sets when analyzing time series data because of the potential high correlation between these data sets. Thus, to invest in real markets, training sets with previous dates sooner than the predicted date of time series data must be defined to successfully hide test data set from the model.
In \cite{qian2017financial}, the authors compared various models, including traditional machine learning techniques such as ARIMA, SVMs, deep learning techniques like DAE (Denoising Autoencoder), and mixture of the above. Their results showed that DAE-SVM produced results superior to those of other methods. Moreover, results were generally more favorable with time series models than simple machine learning models.
In \cite{pyo2017predictability}, the authors predicted the KOSPI 200 index with SVMs and ANNs using Google Trends. Results showed a maximum 52$\%$ accuracy with shorter periods using Google Trends; however, this is not sufficient for real-world investors. 
 The above research can be interpreted as showing that deep learning techniques are more effective than other machine learning techniques in extracting high-level representations of input features, thereby enhancing their overall performance.

In addition, 1D CNNs have shown superior performance for classifying sequential data. A CNN is widely used in image-related tasks such as classification, segmentation, denoising, super-resolution, etc. These days, in addition to image area classification, 1D CNNs are also very effective for deriving important features from segments of whole sequential data and where the location of the features within the segment is insignificant. This applies well to the analysis of sequential sensor data, fixed length periodic signals, and NLP. 
In \cite{gao2014modeling} and \cite{shen2014latent}, semantically meaningful representations of sentences are learned using CNNs in NLP. The models these papers recommend potentially interesting documents to users based on what they are currently reading. 
In \cite{kim2014convolutional}, the authors evaluate a CNN architecture on sentiment analysis and topic categorization tasks. The 1D CNN architecture achieved remarkable results compared with those in previous papers. Also, the network used in this paper is quite simple and easy to implement.

 Time series classification tasks have increasingly been performed with recurrent neural networks in recent years. Moreover, LSTM is widely used in sequential data tasks, such as sequence labelling \cite{kawakami2008supervised}, speech recognition \cite{graves2013speech}, anomaly detection \cite{malhotra2015long}, and financial time series prediction \cite{bao2017deep}.
 Many types of time series problems have used simple or stacked LSTM models for successful predictions. However, despite their advantages for time series data, LSTM models have limitations concerning vanishing gradients and simple feature loss in long sequences. Vaswani \cite{vaswani2017attention} suggested “attention” mechanisms and showed many ways in which they are effective in solving such problems. Attention is a simple vector, and sometimes represents probability distribution using a softmax function. Recurrent neural networks or other deep learning models should take an input as complete sequential data and compress all information into a fixed-length vector as output of the previous model. This implies that even a long data string, represented by fixed-length or final time steps with an output length shorter than the input length, will surely be outputted to information loss. However, attention partially fixes this problem. It allows a model to analyze all the information from the original data and generate the proper output. Attention networks are becoming more widely used for image captioning \cite{vinyals2015show}, neural machine translation \cite{bahdanau2014neural}, and questioning and answering \cite{yu2018qanet}. In \cite{vaswani2017attention}, the authors built multi-head attention modules to replace the recurrent or convolutional neural networks (CNNs) most commonly used in encoder-decoder architectures. This attention model is placed between the encoder and the decoder, and it aids the decoder in selecting the encoded inputs necessary for each step of the decoding process. Results showed superior performance for translation tasks than previous machine translation tasks at a lower computational cost. Finally, this model not only trained faster but also outperformed all previously reported ensembles.

\section{Dataset}
The KOSPI 200 index is a weighted combination of the 200 most traded securities in the Korean stock exchange. In this paper, we take the change ratios of the KOSPI 200, and various indexes such as the currency, commodities, and global indexes, which are closely related with Korean financial markets in terms of fundamental analysis, as input data. We gather our data using the pandas-datareader and calculate the return of a single day as $r[t+1] = \frac {close \; price[t+1] - close \; price[t]}{close \; price[t+1]}$ and initial data $r[0]$ are also given. In addition, we set lookback days as p trading days of a target index. If a particular day is not included in the trading days of a target index, we remove that day and re-calculate the return with a previous close price (using p=10). We write our input as $x[t] = [ R^t[0] ; \cdots ; R^t[p] ]$, where $R^t[i]$ is a collection of input indexes returned on day t. KOSPI trends determine whether or not a target output is larger than 0. Furthermore, we build a binary classification model by analyzing time series data as input. We predict q trading days (using q=19) after the final day of each input. We define trends as
$$trend[t] = \frac {close \; price[t+q] - close \; price[t]}{close \; price[t]}$$
and target labels as

\begin{equation*}
    y[t]=\begin{cases}
        [1, 0], & \text{if trent[t]<0} \\
        [0, 1], & \text{otherwise}.
    \end{cases}
\end{equation*}

that we describe as a one-hot vector.

\section{Methodologies}

 We built multiple deep learning models using multilayer perceptron (MLP), one-dimensional CNNs(1D CNNs), stacked LSTM (stacked LSTM), attention networks, and weighted attention networks. These methods are quite popular for sequential data tasks and show results superior to those of traditional machine learning techniques. In this section, we will describe details of various methods.

\subsection{MLP}

\begin{figure}
\centering
\includegraphics[width=2.5in]{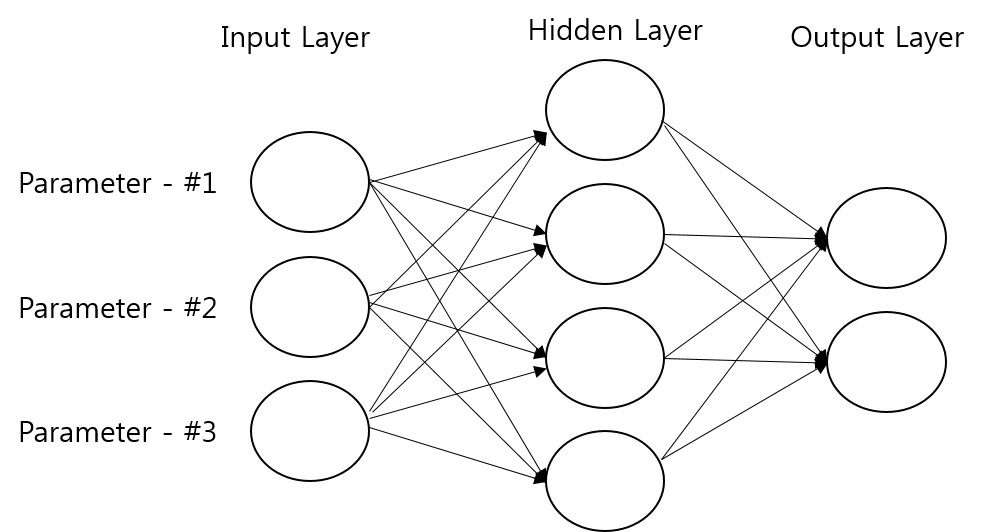}
\caption{An example of a simple MLP with one hidden layer that consists of four hidden neurons}
\label{fig:fig1}
\end{figure}

 MLP consists of a minimum of three fully connected layers. Apart from the input and output layers, each layer contains hidden neurons, which are trainable. Figure \ref{fig:fig1} shows an example of an MLP model with one hidden layer and three neurons. The neurons in the hidden layer take the values of the input parameters, sums them with multiplied assigned weights, and adds a bias. The value of the outputs is determined by applying a transfer function. The number of neurons in the input layer corresponds to the number of input parameters \cite{moghaddam2016stock}.
In \cite{alberg2017improving}, the authors achieved better results for compounded annual returns with an MLP than with a standard factor model. Sezer \cite{sezer2017artificial} predicted Dow 30 stocks with a model trained with data from the daily stock prices between 1997 and 2007, and tested with data from 2007 to 2017. They achieved results comparable to the “buy and hold” strategy in most cases when setting the appropriate technical indicators. 
 We built a simple MLP model with four layers. Each hidden layer contained 64 hidden neurons for our experiments. It is important to note that the input for an MLP model should be a vector; thus, we flattened the input matrix to create an input vector, which was used to feed our model.

\subsection{1D-CNN}

\begin{figure}
\centering
\includegraphics[width=2.5in]{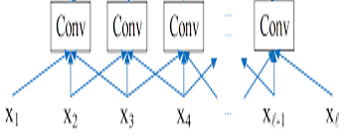}
\caption{An example of a 1D-CNN with sliding kernel size 3}
\label{fig:fig2}
\end{figure}

Many applications of CNNs have focused on image area classification, particularly after ImageNet \cite{krizhevsky2012imagenet} showed superior classification results and accuracy than other image classification methods. However, because of time series property that current state relates with before state, only recurrent neural networks used for analyzing time series analysis. In \cite{kim2014convolutional}, sentence-level classification tasks were performed, which include sentiment analysis and question classification with 1D-CNN, with favorable results.
A 1D CNN is expected to capture data locality well when a kernel slides across the input data. In Figure \ref{fig:fig2}, a sliding kernel with size 3 performs convolution operations over input parameters and produces output for each location. We must choose hyper-parameters like kernel size and number for better results, as the output from 1D CNNs should represent local patterns and various pattern types can be found in various kernel types. One-dimensional CNNs work with patterns in one dimension and tend to be useful in signal analysis over fixed-length signals.
In this paper, we tried several 1D CNN models; we recommend using a model with two convolutional layers and three fully connected layers to predict target labels.

\subsection{LSTM}

 LSTM \cite{hochreiter1997long} is one of the most popular models of recurrent neural networks. These networks are widely applicable for various aspects of sequential data problems, such as speech analysis, sentimental analysis, voice recognition, and financial analysis owing to their particular characteristics. These characteristics can prevent the loss of important features and whole sequences using long-term memory while retaining short-term memory (as with simple recurrent neural networks).
 In \cite{fischer2018deep}, the authors used LSTM networks to predict the movements of the S$\&$P 500 from 1992 to 2015. This LSTM model outperformed memory-free classification methods like random forest classifiers, MLP, and logistic regression classifiers. Furthermore, \cite{heaton2017deep} proposes an LSTM model to forecast stock prices and indexing problems. Deep learning outperformed other classical models owing to its nonlinearity and overcame issues with in-sample approximation quality.

\begin{figure}
\centering
\includegraphics[width=2.5in]{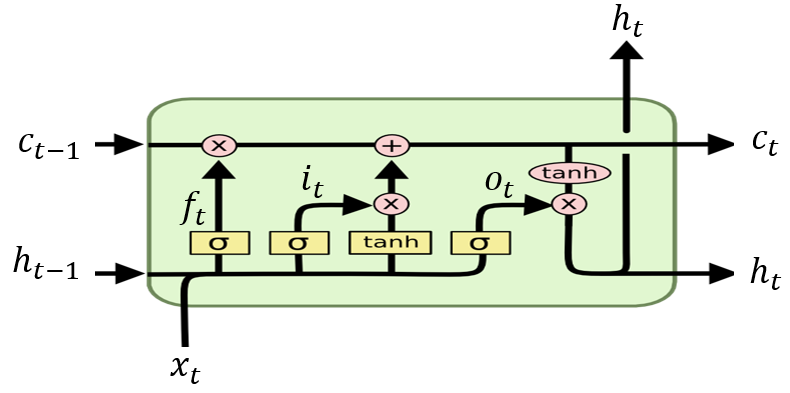}
\caption{An example of an unit of LSTM}
\label{fig:fig3}
\end{figure}
\textbf{}
 One unit of time is shown in Figure \ref{fig:fig3}, and the simple forms of the equations for the front of an LSTM unit with a forget gate are as follows: \\
\begin{align*}
f_t &= \sigma_g(W_{f} x_t + U_{f} h_{t-1} + b_f) \\
i_t &= \sigma_g(W_{i} x_t + U_{i} h_{t-1} + b_i) \\
o_t &= \sigma_g(W_{o} x_t + U_{o} h_{t-1} + b_o) \\
c_t &= f_t \odot c_{t-1} + i_t \odot \sigma_c(W_{c} x_t + U_{c} h_{t-1} + b_c) \\
h_t &= o_t \odot \sigma_h(c_t)
\end{align*}
where $f_t$ is a forget gate, $i_t$ is an input gate, $o_t$ is an output gate, $c_t$ is a cell state, $h_t$ is a hidden state, $\sigma$ is an activation function, and the operator $\odot$ denotes the Hadamard product. We trained the weight matrices as W, U, and b.\

We trained our model with stacked LSTM to understand the more complex features among the data parameters. Stacked LSTM is widely used in sequential data tasks, such as speech recognition \cite{graves2013speech}, anomaly detection \cite{malhotra2015long}, and financial time series prediction \cite{bao2017deep}.

\subsection{Attention Networks}
\begin{figure}
\centering
\includegraphics[width=2.5in]{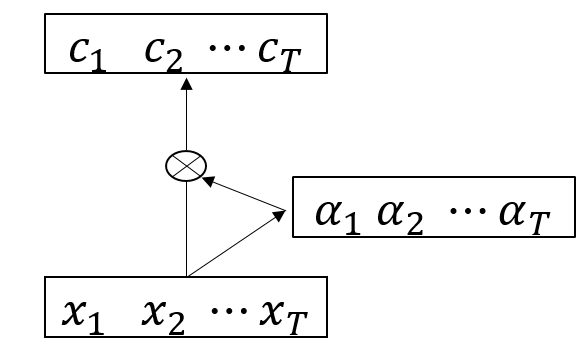}
\caption{Soft attention networks for input vector}
\label{fig:fig4}
\end{figure}

The core function of a sequential data model is to assign the probability of data using a Markov assumption. Owing to the various input lengths, an RNN is naturally introduced to model the conditional probabilities among time points (i.e., a Markov chain model). 
$${ P(w_1 w_2 \cdots w_n ) \approx \prod_{i} P(w_i \mid w_{i-k} \cdots w_{i-1})}$$
In general, vanilla RNNs have common problems with vanishing/exploding gradients and structural problem like ordering. Similar to adding fully connected layers, we set an identical number of hidden neurons with previous input lengths.
The attention model aids in selecting only the inputs of previous layers that are critical for each subsequent step in the model. Once we calculated the importance of each encoded vector, we normalized the vectors using a softmax function and multiplied each encoded vector according to its weight to obtain time-dependent input encoding, which was then fed into each step of the decoder RNN.
There are several types of attention that are dependent on how attention is defined. In our experiment, the alignment weights were learned and placed “softly” all over the input; this was called “soft attention.” However, if one part of the input was selected a time is called, this was called “hard attention.” Soft attention makes the model smooth and differentiable, but expensive when the input is large. Hard attention has a low computational cost but requires more complicated techniques (such as reinforcement learning) to train as the model is non-differentiable. In our experiments, we used soft attention module, as shown in Figure \ref{fig:fig4}, the equations for which are as follows: 
\begin{align*}
& e_{t} = tanh ( W_a [x_1 , x_2 , \cdots , x_T] + b ) \\
& \alpha _ {t} = \frac {exp (e_t )} {\sum _{k=1} ^ T exp(e_k)} 
\end{align*}

In these equations, the attention probabilities $\alpha = (\alpha_1, …, \alpha_T)$ output the softmax function of a vector based on the input sequence $x_t$ and the trainable weights matrix $W_a$. Next, we derived the output of attention as $[c_1 , c_2 , \cdots, c_T ] = [x_1 , x_2 , \cdots, x_T ] * [\alpha_1 , \alpha_2 , \cdots, \alpha_T ]$.

\begin{figure}
\centering
\includegraphics[width=2.5in]{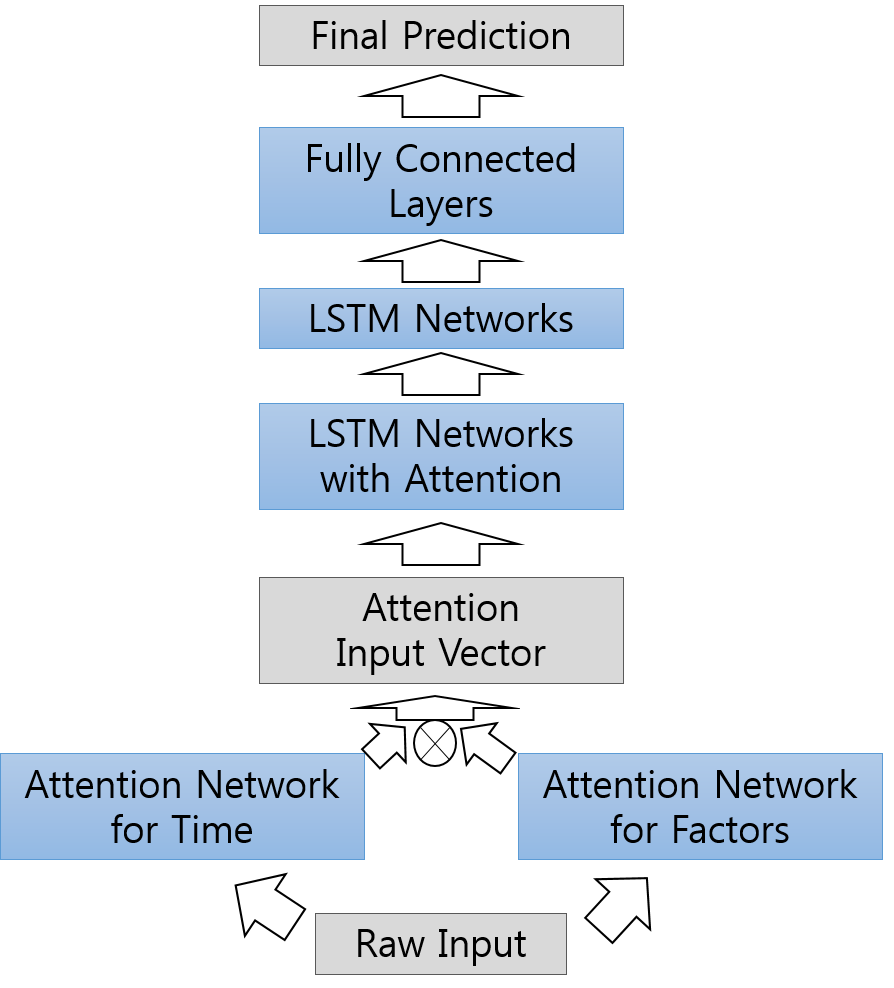}
\caption{Attention networks for KOSPI200 prediction}
\label{fig:fig5}
\end{figure}

This vector intuitively summarizes the importance of the different elements in the input. Thus, we visualized this by plotting an attention vector as a bar.
We tested various configurations, such as attention before LSTM, after LSTM, along with attention for time aspects, factor aspects, or both. For our final results, we used attention networks with both time and factor aspects, then multiplied the attention vectors to produce final vectors as an input data for the LSTM networks. Before moving to the next LSTM network layer, we derived output with the attention networks using following equations: 

\begin{align*}
& e_{j, t} = V_a \cdot tanh ( W_a s_{t-1} + U_a h_j ) \\
& \alpha _ {j, t} = \frac {exp (e_j )} {\sum _{k=1} ^ T exp(e_k)} 
\end{align*}

where the input sequence $h_j$, the internal hidden state of the output cell $s_{t-1}$, and the trainable matrix $V_a$ appear.
After applying attention networks with LSTM, we ran LSTM again as a stacked LSTM network and then passed through the fully connected layers to derive final predictions. Our full model architecture is shown in Figure \ref{fig:fig5}.

\subsection{Weighted Attention Networks}

After building the attention network model (as described above), we tested a model using weighted attention networks with a modified loss function. As we trained various models with categorical cross entropy as loss function, and labels were defined with one-hot vectors, our trained models were focused only on hit ratios. However, in financial markets, it is essential to consider the change ratio over a day. Thus, we sought accurate predictions on a day where large changes occurred. We customized our loss functions by multiplying the absolute values of the change ratio of the test data and original categorical cross entropy values. 
 Categorical cross entropy functions for the binary classes were defined as 
$$H_{y'}(y) := - \sum_{i} ({y_i' \log(y_i) + (1-y_i') \log (1-y_i)})$$,
where $y_i$ is the predicted probability for class i and $y_i'$ is the true probability. Our weighted categorical cross entropy was defined as 
$$H^{weighted}_{y'}(y) := - abs(change\_ratio_i) * \sum_{i} ({y_i' \log(y_i) + (1-y_i') \log (1-y_i)})$$
where $change\_ratio_i$ is the change ratio of the test data i.
Therefore, our model updated to ensure accurate predictions for large change ratios and to minimize loss.

\section{Experimental Results}

 We trained and tested our models using various deep learning models, as mentioned in Section 4. 
 Furthermore, we defined our measurement criteria as hit ratios, which are the precision of trends prediction. These definitions were formulated as follows:
$$ hit \; ratio = \frac{ \sum_{i=1} ^N prediction_i } {N} , $$
and 
\[
prediction_i = 
\begin{cases}
 1 & if \; prediction_i \cdot real_i > 0 \\ 
 0 & otherwise 
\end{cases}
\]
where $prediction_i$ denotes the prediction of the i-th sample change ratio, and $real_i$ denotes the real market change ratio.

We set our training data to precede the test data; the test data will be unseen in our model. Therefore, we subsampled our training data set to use validation sets with a split ratio of 0.7 ($70\%$ for training and $30\%$ for validation, randomly over the entire data set.) We set our training data from the first trading day of 2000 until the end of 2016, and our test data from the first trading day of 2017 to the end of July 2018. We choose the best model with highest hit ratio and then tested the target data. 

\subsection{Best Lookback Days}

First, we conducted experiments to determine the number of lookback necessary to ensure optimal accuracy for the proposed deep learning models. Next, we set our lookback days to 5, 10, 15, 20, 30, and 60 days; five trading days are regarded as one week. Thus, we used 1, 2, and 3 lookback weeks, and 1 and 2 months for our input data. For the 5 and 10 lookback day periods, we tested days from the day after a lookback day to 5 and 10 days after a lookback day. In addition, for 15, 20, 30, and 60 days, we predicted for the day after lookback days to 1 week, 2 weeks, and so on until the predicted days reached the same number of days as the lookback period. For example, if the lookback period was 15 days and the prediction period was 10 days, our input data comprised information from a 3-week period and we predicted a trend for the 2 weeks following the last day of the lookback period. 

\begin{figure}
\centering
\includegraphics[width=2.5in]{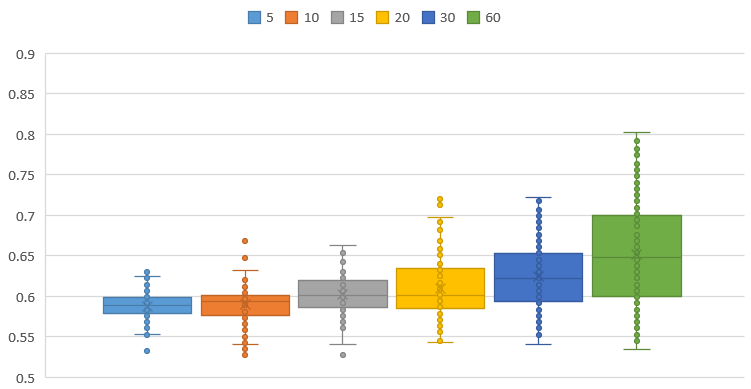}
\caption{Hit ratio with respect to lookback days}
\label{fig:fig6}
\end{figure}

In Figure \ref{fig:fig6}, each bar from right to left implies hit ratios of 5, 10, 15, 20, 30, and 60 lookback days. These results represent the best hit ratios for the test data with various deep learning models. We do not use the best model for validation data because we want to find the best lookback days. As shown in Figure \ref{fig:fig6}, longer lookback days produced better hit ratios, particularly over 60 lookback days. Our best model for test data approached approximately 80$\%$; a high hit ratio in financial prediction.
Therefore, in next subsection, we try to compare various deep learning models with 60 lookback days and various prediction days.

\subsection{Results of Various Deep Learning Models}

\begin{figure}
\centering
\includegraphics[width=2.5in]{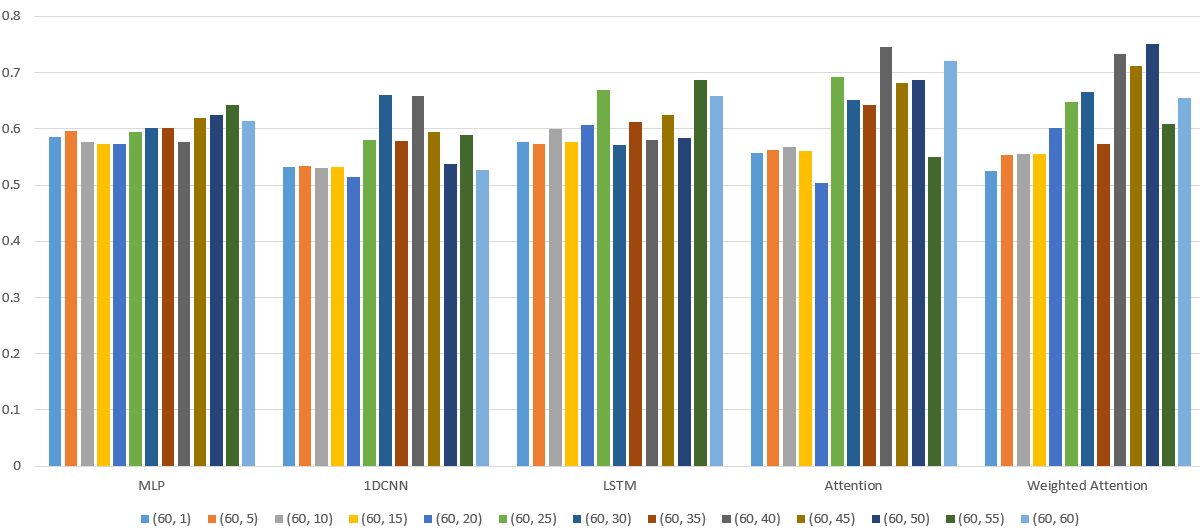}
\caption{Hit ratio with respect to prediction days with various deep learning models}
\label{fig:fig7}
\end{figure}

As previously mentioned, we tested various deep learning models with 60 lookback days and prediction days from the following day to every week until prediction days reached 60. In this experiment, we produced hit ratios  to determine which model was superior without seeing the test data. We trained each model five times to verify their stability; each model returned similar hit ratio results. Figure \ref{fig:fig7} represents the average of hit ratio of each models with various prediction days.

 As shown in Figure \ref{fig:fig7}, attention networks outperform other models over 60 lookback days. For long-term time series prediction, we assumed that LSTM brings advantages owing to its memory cell properties. In addition, LSTM shows superior hit ratios compared with MLP and 1D CNN, which are simplistic methods. In \cite{pyo2017predictability}, the best hit ratio results peaked at 0.52 when using ANN and SVMs with or without Google Trends (we assumed our MLP and ANN models in paper \cite{pyo2017predictability} were quite similar, as their ANN accuracy of 0.51 is almost identical to our MLP results).
 With both attention models (weighted or unweighted), we produced optimal results when predicting 40 days after the data input ended. In addition, we produced a hit ratio of 0.715 using attention networks with the test data set and 0.763 using weighted attention networks with our best validation models. Moreover, we analyzed these models further using positive and negative trends and by accounting for the occurrence of comparably large change ratios. 

\begin{table}[h!]
\begin{center}
 \begin{tabular}{| c | c | c | c |} 
 \hline
  & Dataset & Attention Networks & Weighted Attention Networks \\ 
 \hline
 Positive & 0.602 & 0.709 & 0.825 \\ 
 \hline
 Negative & 0.398 & 0.723 & 0.671 \\
 \hline
 Total & 1.0 & 0.715 & 0.763 \\
 \hline
 Earn  points & 1525.20 & 1090.718 & 1257.462 \\
 \hline
\end{tabular}
\caption{Best Model Results}
\label{table:1}
\end{center}
\end{table}

As shown in table \ref{table:1}, our test data set showed positive and negative trends at a ratio of 6:4. If we corrected all up and down trends, we earned 1525.20 points. Furthermore accuracy improved with weighted attention networks at positive trends and attention networks at negative trends. Thus, we earned 15$\%$ more points than plain attention networks when using weighted attention networks.

\subsection{Visualization Attention Vectors}
 In section 4.4, we mentioned attention networks bringing great advantages for visualizing attention vectors to analyze which parameters our model considered carefully using a greater weight.
If we changed the input data, the attention vectors changed in tandem. Thus, we present just one example of attention vector visualization when using input data with the highest change ratio.

\begin{figure}
\centering
\includegraphics[width=2.5in]{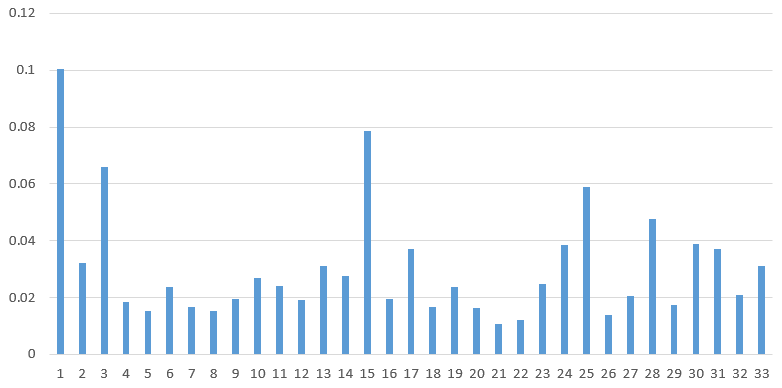}
\caption{Attention vector of factors}
\label{fig:fig8}
\end{figure}

\begin{figure}
\centering
\includegraphics[width=2.5in]{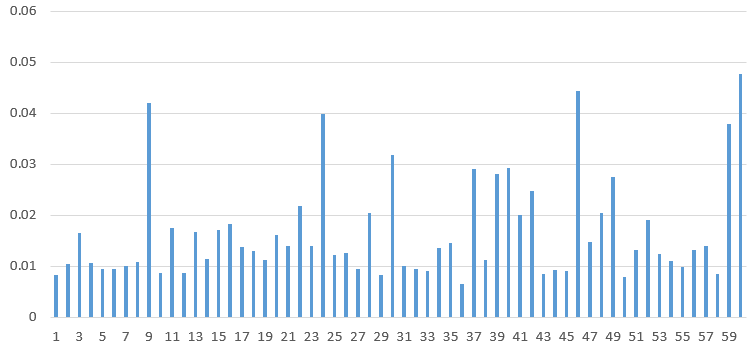}
\caption{Attention vector of times}
\label{fig:fig9}
\end{figure}

We used two types of attention vectors in our attention models. Figure \ref{fig:fig8}, and Figure \ref{fig:fig9} represent the attention vectors for factors and times, respectively, when using the input data with highest change ratios among our test data with weighted attention networks. In this case, with respect to factors, the dollar currency index and S$\&$P 500 global index were the highest weighted factors. With respect to time, the last day of input data and a few days in the middle had the highest weights. From the perspective of time, our results were affected most by the last day of the lookback days and middle days in this example. However, only 4$\%$ of whole weights represented the largest value. It is still difficult to analyze these vectors in detail as they have complex relations and pass through our model to determine whether a trend is rising or falling. However, we can determine which is the most effective factor and time intuitively through attention vector visualization.

\section{Conclusion}

In this study, we tested various deep learning models for predicting the trends of the KOSPI 200 index. In particular, we tested MLP, 1D CNN, LSTM, and attention networks, which are widely used in sequential data applications. While short lookback days showed low hit ratios with various models, long lookback days with 60 trading days returned higher hit ratios with various models. This implies that including more days with input data provides better hit ration accuracy. When considering 60 trading days as lookback days, using 40 trading days as prediction days returned the highest hit ratio with the attention networks model. Moreover, the highest earn points were returned when we used weighted attention networks, as loss functions were minimized when improved at higher change ratios. 
Through our experimental results, we can confirm with several findings. First, LSTM works well with sequential data, which depends more on time than MLP and 1D CNN. Second, longer lookback days produced a higher probability of better hit ratios and better overall results with the models that include LSTM networks. Finally, weighted attention networks perform better with long sequential data and have the advantage of visualization for intuitively analyzing models.

\bibliographystyle{unsrt}

\end{document}